\documentclass[10pt,twocolumn,letterpaper]{article}

\usepackage{cvpr}              % To produce the CAMERA-READY version
\usepackage[dvipsnames]{xcolor}

\definecolor{cvprblue}{rgb}{0.21,0.49,0.74}
\usepackage[pagebackref,breaklinks,colorlinks,citecolor=cvprblue]{hyperref}
\usepackage{booktabs}
\usepackage{multirow}
\usepackage{pifont}% http://ctan.org/pkg/pifont
\newcommand{\cmark}{\ding{51}}%
\usepackage{lipsum}

\newcommand\blfootnote[1]{%
  \begingroup
  \renewcommand\thefootnote{}\footnote{#1}%
  \addtocounter{footnote}{-1}%
  \endgroup
}

\def\paperID{2311} % *** Enter the Paper ID here
\def\confName{CVPR}
\def\confYear{2025}

\title{PICD: Versatile Perceptual Image Compression with Diffusion Rendering}

\author{Tongda Xu$^{1,2,*}$, Jiahao Li$^2$, Bin Li$^2$, Yan Wang$^1$, Ya-Qin Zhang$^1$, Yan Lu$^2$\\
$^1$AIR, Tsinghua University, $^2$Microsoft Research Asia\\
{\tt\small \{xutongda,wangyan,zhangyaqin\}@air.tsinghua.edu.cn,\{li.jiahao, libin, yanlu\}@microsoft.com}
}

\begin{document}
\twocolumn[{
\maketitle
\begin{center}
    \captionsetup{type=figure}
    \includegraphics[width=\textwidth]{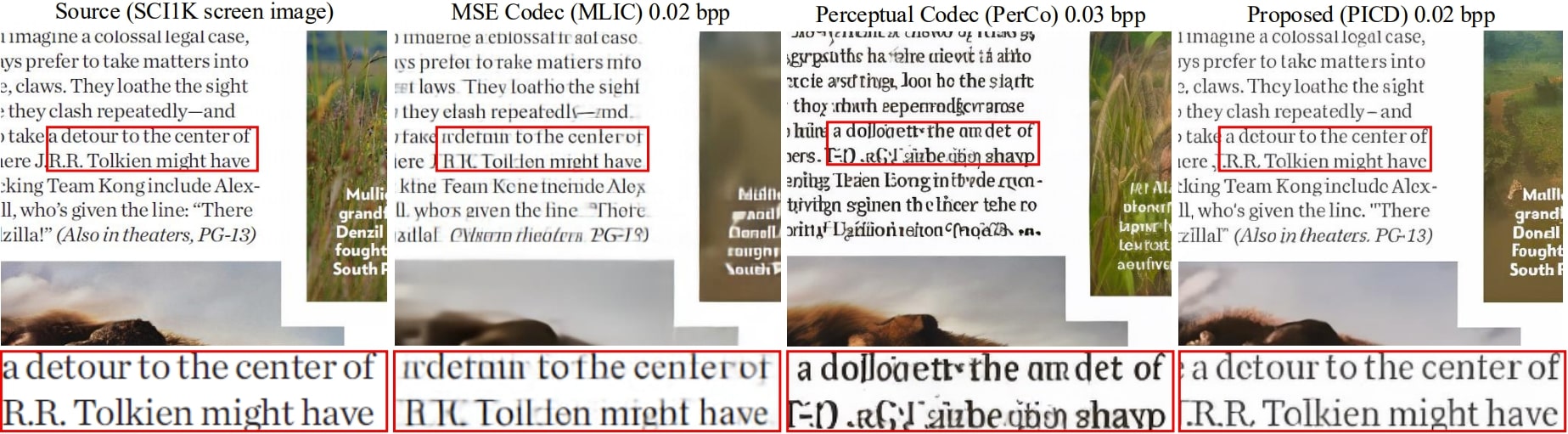}
    \includegraphics[width=\textwidth]{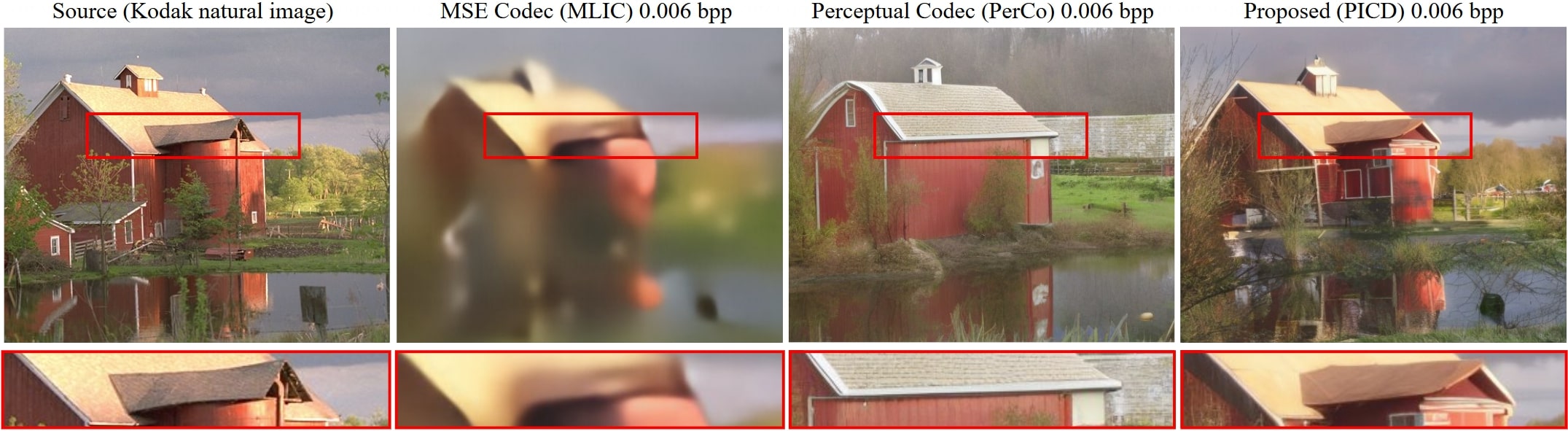}
    \captionof{figure}{For both screen and natural images, PICD demonstrates high text accuracy and superior visual quality simultaneously.}
    \label{fig:cov}
\end{center}
}]
\begin{abstract}

Recently, perceptual image compression has achieved significant advancements, delivering high visual quality at low bitrates for natural images. However, for screen content, existing methods often produce noticeable artifacts when compressing text. To tackle this challenge, we propose versatile perceptual screen image compression with diffusion rendering (\textbf{PICD}), a codec that works well for both screen and natural images. More specifically, we propose a compression framework that encodes the text and image separately, and renders them into one image using diffusion model. For this diffusion rendering, we integrate conditional information into diffusion models at three distinct levels: 1). Domain level: We fine-tune the base diffusion model using text content prompts with screen content. 2). Adaptor level: We develop an efficient adaptor to control the diffusion model using compressed image and text as input. 3). Instance level: We apply instance-wise guidance to further enhance the decoding process. Empirically, our PICD surpasses existing perceptual codecs in terms of both text accuracy and perceptual quality. Additionally, without text conditions, our approach serves effectively as a perceptual codec for natural images.
\blfootnote{$^*$The work was done when Tongda Xu was a full-time intern with Microsoft Research Asia.} 
\end{abstract} 

\begin{figure}[htb]
  \centering
   \includegraphics[width=\linewidth]{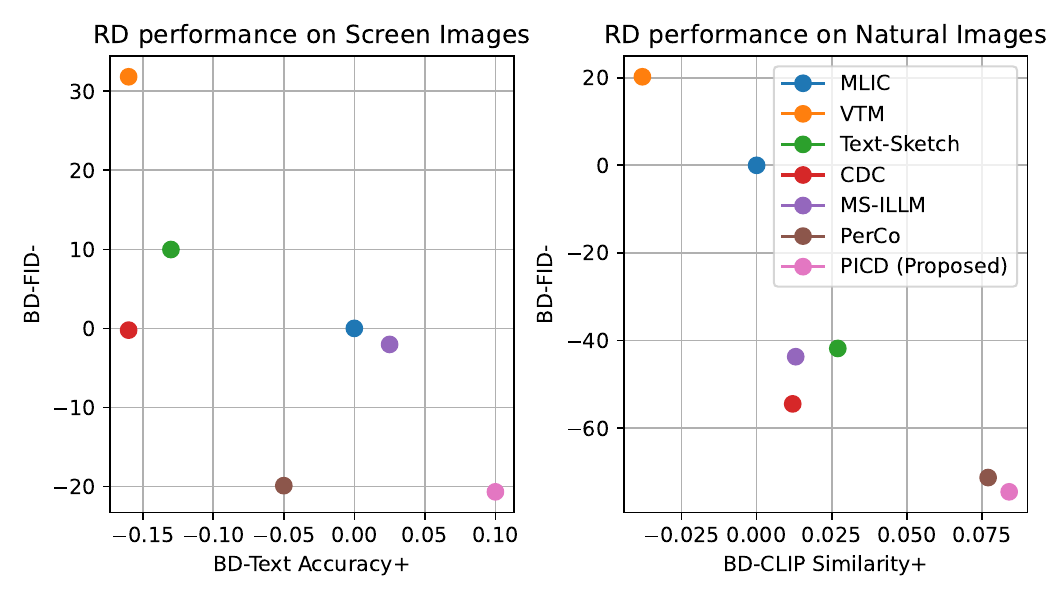}
   \caption{PICD works well for both screen and natural images.}
   \label{fig:covrd}
\end{figure}
\section{Introduction}

With recent advancements in generative models, perceptual image compression has acquired the capability to compress natural images at low bitrate while maintaining high visual quality  \citep{muckley2023improving,Careil2023TowardsIC}. These approaches are driven by the rate-distortion-perception trade-off \citep{blau2018perception,blau2019rethinking} and typically involve training a conditional generative model as a decoder, such as a generative adversarial network (GAN) \citep{Rippel2017RealTimeAI,tschannen2018deep,Mentzer2020HighFidelityGI,Agustsson2022MultiRealismIC,muckley2023improving,Jia2024GenerativeLC} or a diffusion model \citep{yang2022lossy,hoogeboom2023high,Careil2023TowardsIC,Lei2023TextS,Ma2024CorrectingDP,Relic2024LossyIC}.

Although these perceptual codecs are effective for natural images, they generally fall short for screen images. This limitation arises because current perceptual codecs focus on preserving marginal distributions rather than accurately reproducing text. For example, a screenshot containing the character "a" might be decoded as the character "c." As long as the character "c" is clear and visually coherent, the codec is deemed to be perceptually lossless. However, such reconstruction is obviously unacceptable for screen content. Conversely, existing screen content codecs prioritize text accuracy but disregard perceptual quality. \citep{Mitrica2019VeryLB,Tang2022TSASCCTS,Och2023ImprovedSC,Heris2023MultiTaskLF,zhou2024enhanced,Lai2024LearnedIC}. They enhance text quality based on non-perceptual natural codec, and produce blurry reconstructions at low bitrates.

To address the aforementioned challenges, we propose versatile perceptual screen content codec with diffusion rendering (PICD) for both screen and natural images. Specifically, we encode the text information losslessly and render them with a compressed image using a diffusion model to achieve high text accuracy and visual quality. To implement this diffusion rendering, we introduce a three-tiered conditioning approach: 1) Domain level: We fine-tune the base diffusion model using text content prompts. 2) Adaptor level: We develop an efficient adaptor that is conditioned on both text content and its corresponding location, in conjunction with compressed image. 3) Instance level: We employ instance-wise guidance during decoding to further enhance performance. Empirical results demonstrate that our proposed PICD surpasses previous perceptual codecs in terms of both text accuracy and perceptual quality. Furthermore, our PICD remains effective for natural images.

\begin{itemize}
    \item We propose PICD, a versatile perceptual codec for both natural and screen content.
    \item We introduce a highly efficient conditional framework to transform a pre-trained diffusion model into a codec, with domain, adaptor, and instance level conditioning.
    \item We demonstrate that PICD achieves both high visual quality and text accuracy for screen and natural images.
\end{itemize}

\section{Preliminary: Perceptual Image Codec}
Perceptual image compression mains to maintain the visual quality of the decoded image. Specifically, denoted as $X$, the encoder as $f_{\theta}(\cdot)$, the bitstream as $Y = f_{\theta}(X)$, the decoder as $g_{\theta}(\cdot)$, and the reconstructed image as $\hat{X} = g_{\theta}(Y)$, driven by rate-distortion-perception trade-off \citep{agustsson2019generative,Mentzer2020HighFidelityGI,muckley2023improving}, perceptual image compression constrains the marginal distribution of $\hat{X}$ to match that of the source $X$:
\begin{gather}
    p(\hat{X}) = p(X).
\end{gather}
The majority of perceptual image codecs use a conditional generative model as the decoder \citep{agustsson2019generative,Mentzer2020HighFidelityGI,muckley2023improving,hoogeboom2023high,Careil2023TowardsIC}. Specifically, these models train the decoder $g_{\theta}(\cdot)$ to approximate the true posterior distribution $p(X|Y)$:
\begin{gather}
    \hat{X} = g_{\theta}(Y) \sim p(X|Y).
\end{gather}
In this scenario, the marginal distribution $p(\hat{X})$ aligns with $p(X)$, thereby achieving the optimum perceptual quality.

% \textbf{Denoising Diffusion Probability Model} Diffusion models are category of generative models using a $T$-step Gaussian Markov chain \citep{sohl2015deep}. Among these, the Denoising Diffusion Probability Model (DDPM) is one of the most commonly adopted frameworks \citep{ho2020denoising}. Let us denote the source image as $X_0$. The forward Markov process in DDPM is expressed as:
% \begin{gather}
%     q(X_T,...,X_1|X_0) = \prod_{t=1}^T q(X_t|X_{t-1}), \notag \\ \text{where } q(X_t|X_{t-1}) = \mathcal{N}(\sqrt{1-\beta_t}X_{t-1}, \beta_tI),\label{eq:dfw}
% \end{gather}
% where $\beta_t$ are predefined hyperparameters. The reverse process in the DDPM is also structured as a Markov chain, with the transition kernel $p_{\theta}(X_{t-1}|X_t)$ being dependent on the score function $\nabla_{X_t} \log p(X_t)$, which is approximated using a neural network $s_{\theta}(X_t, t)$, parameterized by $\theta$:
% \begin{gather}
%     p_{\theta}(X_0,...,X_T) = p(X_T)\prod_{t=1}^T p_{\theta}(X_{t-1}|X_t), \notag \\  \text{where } p_{\theta}(X_{t-1}|X_t) = \mathcal{N}\left(\frac{1}{\sqrt{\alpha_t}}(X_t + \beta_t s_{\theta}(X_t,t)), \sigma^2_t I\right), \label{eq:ddpm2}
% \end{gather}
% with $\alpha_t$ and $\sigma_t^2$ being parameters determined by $\beta_t$ \citep{ho2020denoising}.

\section{PICD: Versatile Perceptual Image Compression with Diffusion Rendering}
\begin{figure*}[thb]
  \centering
   \includegraphics[width=\linewidth]{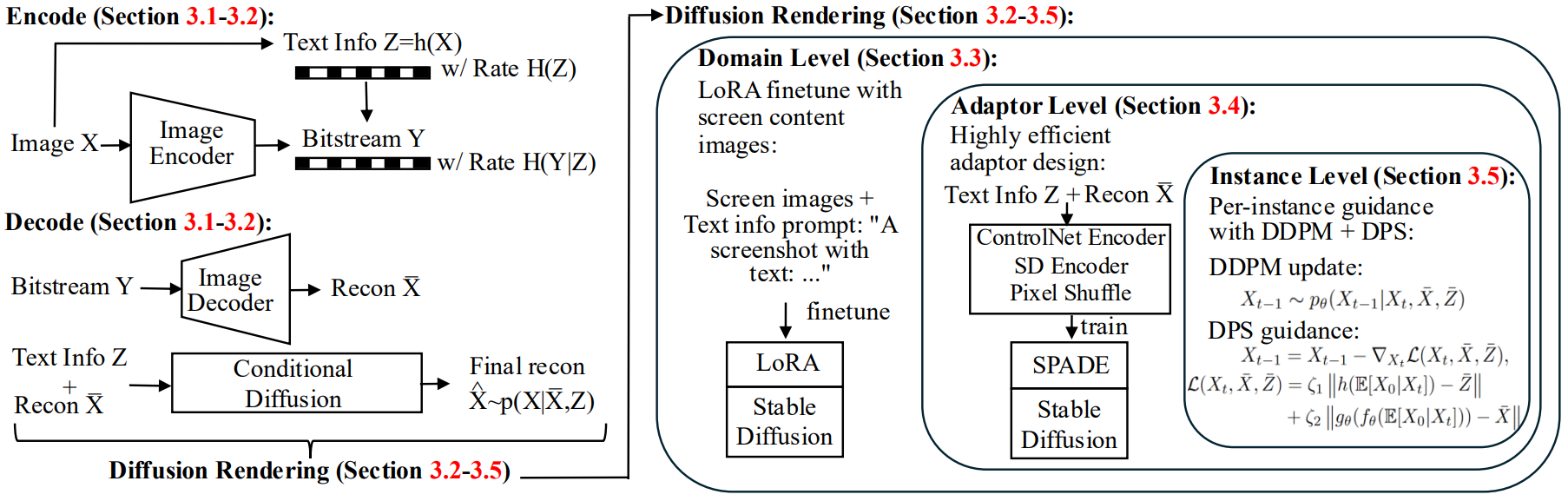}
   \caption{The overall framework of our proposed PICD.}
   \label{fig:van}
\end{figure*}

\subsection{Overall Framework and Rationale}
We first describe the PICD for screen images and then simplify PICD for natural images. The foundational framework of PICD for screen images is as follows:
\begin{itemize}
    \item \textbf{Encoding}: We first extract the text $Z = h(X)$ using Optical Character Recognition (OCR) model $h(.)$ and compress it losslessly. Then, we encode the image $X$ with $Z$ as a condition, into the bitstream $Y = f_{\theta}(X|Z)$.
    \item \textbf{Decoding}: We first decode text $Z$ then the image $\bar{X}=g_{\theta}(Y|Z)$. Next, we merge the $\bar{X}$ with $Z$ into one image using conditional diffusion model $\hat{X}\sim p_{\theta}(X|\bar{X},Z)$. We name this merging process Diffusion Rendering.
\end{itemize}

In the rest of this section, we provide an explanation of the optimality of PICD. Specifically, we show that: 1) PICD is optimal for preserving text information $Z$; 2) PICD achieves best perceptual quality.

\textbf{PICD is Optimal for Preserving Text Information} Let $h(.)$ denote an OCR algorithm extracting text content and its corresponding location, such that $Z = h(X)$ represents the OCR output. We define the reconstructed text as accurate when the OCR output for the reconstructed image, $\hat{Z} = h(\hat{X})$, corresponds exactly to $Z$:
\begin{gather}
    \hat{Z} = h(\hat{X}) = h(g_{\theta}(Y)) = Z. \label{eq:zcond}
\end{gather}
At first glance, one might propose attaining this objective by integrating an additional loss function, such as $||h(\hat{X}) - Z||^2$, into the training process \citep{Lai2024LearnedIC}. However, this approach proves to be ineffective at low bitrates (See Table~\ref{tab:abl2}).

In this study, we introduce a more straightforward approach to tackle this challenge. We begin by encoding $Z$ losslessly, followed by encoding $X$ with $Z$ as a condition. Subsequently, we render $Z$ together with the compressed $\bar{X}$ to produce a single decoded image, $\hat{X}$. The rendering is facilitated by a conditional diffusion model, which employs both the compressed image $\bar{X}$ and the text $Z$, i.e.,
\begin{gather}
    \bar{X} = g_{\theta}(Y|Z),
    \hat{X} \sim p(X|\bar{X},Z).
\end{gather}
As the proposed approach works well for both screen and natural images using diffusion rendering, we name it versatile perceptual image compression using diffusion rendering (PICD). To comprehend why PICD achieves optimal bitrate for preserving text information, we observe that $Z$ should be completely determined by $Y$ as per Eq.~\ref{eq:zcond}. This condition implies that the entropy of $Z$ given $Y$ is zero, i.e.,
\begin{gather}
    H(Z|Y) = 0.
\end{gather}
Then according to Kraft's inequality \citep{cover1999elements}, the optimal bitrate for compressing $Z$ losslessly and then compressing $Y$ given $Z$, is equivalent to compressing $Y$ alone:
\begin{gather}
    H(Y|Z) + H(Z) = H(Y) + H(Z|Y) = H(Y).
\end{gather}    
From this intuitive reasoning, we can conclude that PICD is nearly optimal for preserving text information $Z$.

\textbf{PICD is Optimal for Perceptual Quality} On the other hand, PICD satisfies the perfect perceptual codec constraint $p(\hat{X}) = p(X)$. More specifically, we have:
\begin{gather}
p(\hat{X}) = \int p(\bar{X},Z) p(X|\bar{X},Z) d \bar{X} dZ = p(X).
\end{gather}
Therefore, PICD achieves optimal perceptual quality defined by \citet{blau2018perception}.

\subsection{A Basic Implementation of PICD}
\label{sec:vanilla}
Now, let us consider a basic implementation of the PICD, with a pre-trained image codec and ControlNet \citep{Zhang2023AddingCC}. 

\textbf{Encoder: Text Information Compression} In alignment with previous studies in screen content compression \citep{Tang2022TSASCCTS}, we utilize the Tesseract OCR engine \citep{Smith2007AnOO} as $h(.)$ to extract text information from screen images. The OCR output contains the textual content and three coordinates, which denote the upper-left corner of the word and its height. We concatenate all extracted words into a single string, which is then compressed losslessly using cmix \citep{Knoll2012AML}. The coordinates are compressed exponential-Golomb coding. 
\begin{figure*}[thb]
  \centering
   \includegraphics[width=\linewidth]{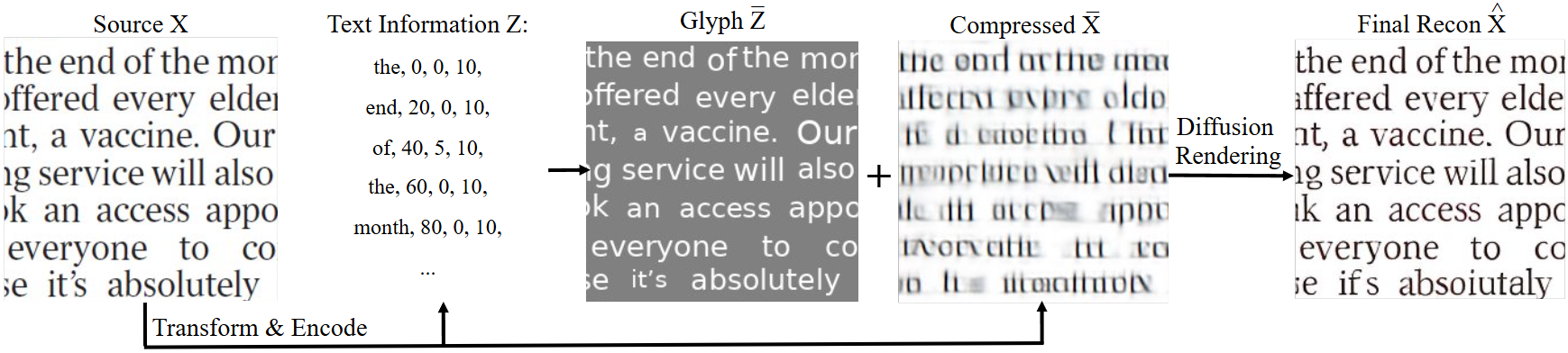}
   \caption{An example of the PICD pipeline. PICD first extracts and encodes text information $Z$ from source $X$. Then, PICD converts text information into text glyph $\bar{Z}$. Finally, PICD renders glyph $\bar{Z}$ with a compressed image  $\bar{X}$ into reconstruction $\hat{X}$ using diffusion.}
   \label{fig:glyph}
\end{figure*}

\textbf{Encoder: Image Compression} For the image compression of $X$, we employ MLIC \citep{jiang2023mlic}, a state-of-the-art image codec. To integrate the text condition $Z$ into MLIC, we first convert $Z$ into an image based on its textual content and placement, referred to as a glyph image, denoted as $\bar{Z}$ (see Figure~\ref{fig:glyph}), which is widely used in diffusion-based text generation tasks \citep{Yang2023GlyphControlGC}. To compress $X$ conditioned on $Z$, we adopt a fine-tuning strategy for a pre-trained MLIC model, similar to the method used in ControlNet \citep{Zhang2023AddingCC}. Specifically, we create a duplicate branch of the MLIC encoder to process the glyph image $\bar{Z}$. The output from this duplicated encoder passes through zero-convolution layers and is subsequently integrated with the original MLIC encoder and decoder. During training, the pre-trained MLIC model is initialized, and the duplicated encoder along with the zero-convolution layers are fine-tuned. (See Appendix A.1).

\textbf{Decoder: Diffusion Rendering} To learn the posterior distribution $p(X|\bar{X},Z)$, we build upon Stable Diffusion, a pre-trained text-to-image generation model. While Stable Diffusion facilitates image generation from text prompts, it does not support specifying text locations or accepting an image as input. Therefore, we integrate a ControlNet \citep{Zhang2023AddingCC} into the Stable Diffusion framework. This ControlNet is designed to process two inputs: the image $\bar{X}$ decoded from MLIC, and the glyph image $\bar{Z}$. For the text prompt, we concatenate the text content from $Z$ and prepend it with the prefix "a screenshot with text: ...". Consistent with prior studies \citep{hoogeboom2023high}, we utilize the decoded image $\bar{Z}$ rather than the bitstream $Y$ as input to the diffusion model.
% \begin{figure}[thb]
%   \centering
%    \includegraphics[width=\linewidth]{fig_van_eg.PNG}
%    \caption{An example output of vanilla implementation and PICD.}
%    \label{fig:glyph}
% \end{figure}

\textbf{Three Level Improvements} Currently, we have obtained a functional perceptual codec. However, this basic implementation has significant limitations. As shown in Figure~\ref{fig:abl} and Table~\ref{tab:abl} (b), its reconstruction suffers from colour drifting and low text quality. Therefore, we propose improvements across three dimensions: \textbf{Domain level}: In Section~\ref{sec:global}, we enhance the base Stable Diffusion model by fine-tuning it with text prompts and screen images. \textbf{Adaptor level}: In Section~\ref{sec:adaptor}, we explore more efficient adaptor mechanisms beyond ControlNet \citep{Zhang2023AddingCC}. \textbf{Instance level}: In Section~\ref{sec:ins}, we incorporate instance-level guidance for further performance optimization.

\begin{table*}[thb]
\includegraphics[width=\linewidth]{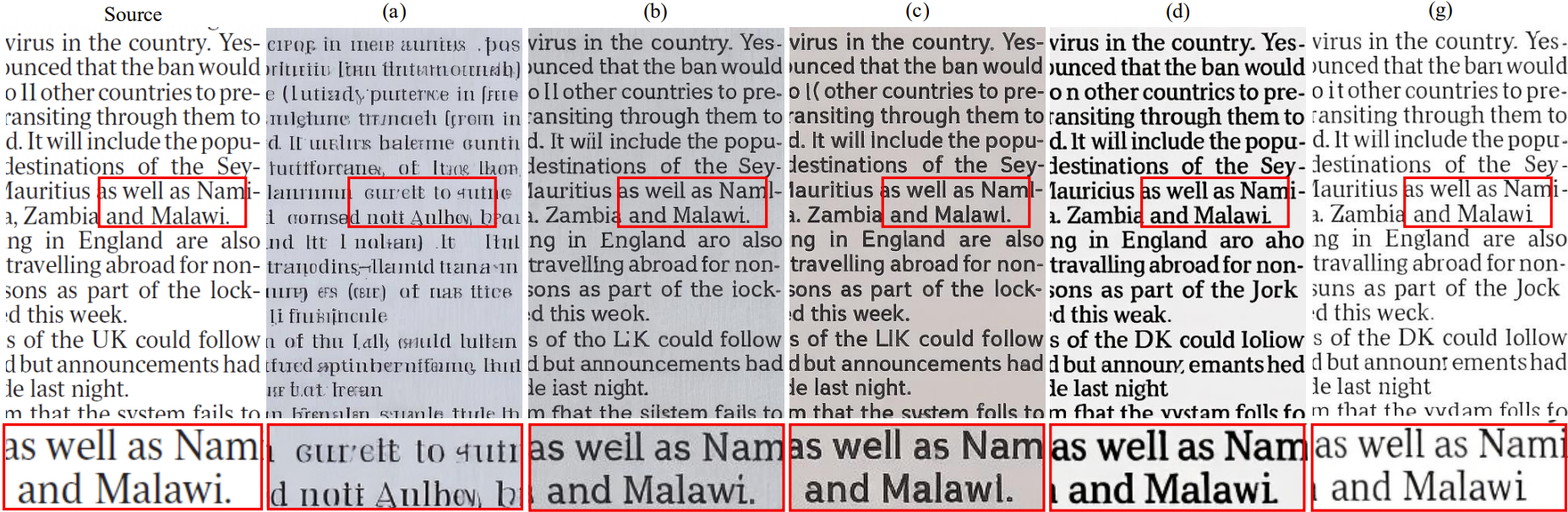}
\captionof{figure}{Ablation studies on different components of diffusion rendering.}
\label{fig:abl}
\centering
\resizebox{\linewidth}{!}{
\begin{tabular}{@{}ccccccccccccc@{}}
\toprule
\multirow{2}{*}{ID} & \multirow{2}{*}{Glyph (Sec 3.2)} & \multirow{2}{*}{Domain Level (Sec 3.3)} & \multicolumn{3}{c}{Adaptor Level (Sec 3.4)} & \multirow{2}{*}{Instance Level (Sec 3.5)} & \multirow{2}{*}{Text Acc$\uparrow$} & \multirow{2}{*}{PSNR$\uparrow$} & \multirow{2}{*}{FID$\downarrow$} & \multirow{2}{*}{CLIP$\uparrow$} & \multirow{2}{*}{LPIPS$\downarrow$} \\ \cmidrule(lr){4-6}
                             & & & ControlNet \citep{Zhang2023AddingCC} & StableSR \citep{Wang2023ExploitingDP} & Proposed &                                    &  &  &  &                       &                        \\ \midrule
                             (a) & & & \cmark &  & & & 0.3468 & 19.10 & 45.83 & 0.8209 & 0.1694 \\
                             (b) &  \cmark & & \cmark & & & & 0.4404 & 18.84 & 45.35 & 0.8617 & 0.1646 \\
                             (c) & \cmark &  & & \cmark & & & 0.3934 & 20.56 & 49.76 & 0.8850 & 0.1344 \\
                             (d) & \cmark &  & & &  \cmark & & 0.4081 & 19.88 & 37.90 & 0.8922 & 0.1376 \\
                             (e) & \cmark &  & \cmark& & & \cmark & 0.4446 & 23.30 & 39.81 & 0.8917 & 0.1225 \\
                             (f) & \cmark &  &  & & \cmark & \cmark & 0.4445 & \textbf{23.70} & 35.54 & 0.9059 &  0.1172 \\
(g) & \cmark & \cmark &  &  & \cmark & \cmark &  \textbf{0.4568} & 23.67 & \textbf{34.77} & \textbf{0.9082} & \textbf{0.1168} \\ \bottomrule
\end{tabular}
}
\captionof{table}{Ablation studies on different components of diffusion rendering.}
\label{tab:abl}
\end{table*}
\subsection{Diffusion Rendering: Domain Level}
\label{sec:global}
Stable Diffusion is trained on natural images and lacks exposure to screen. Consequently, a straightforward enhancement to the basic implementation involves finetuning Stable Diffusion with screen content $X$ and text information $Z$ as prompt. Specifically, we utilize the WebUI dataset \citep{Wu2023WebUIAD}, consisting of 400,000 web screenshots. We employ the Tesseract OCR engine \citep{Smith2007AnOO} to extract text content, which is then concatenated into prompts as described in Section~\ref{sec:vanilla}. To enhance finetuning efficiency, we implement Low-Rank Adaptation (LoRA) \citep{Hu2021LoRALA}, using a rank of 256 instead of full parameter tuning. (See Appendix B.1)

\begin{figure}[thb]
  \centering
   \includegraphics[width=\linewidth]{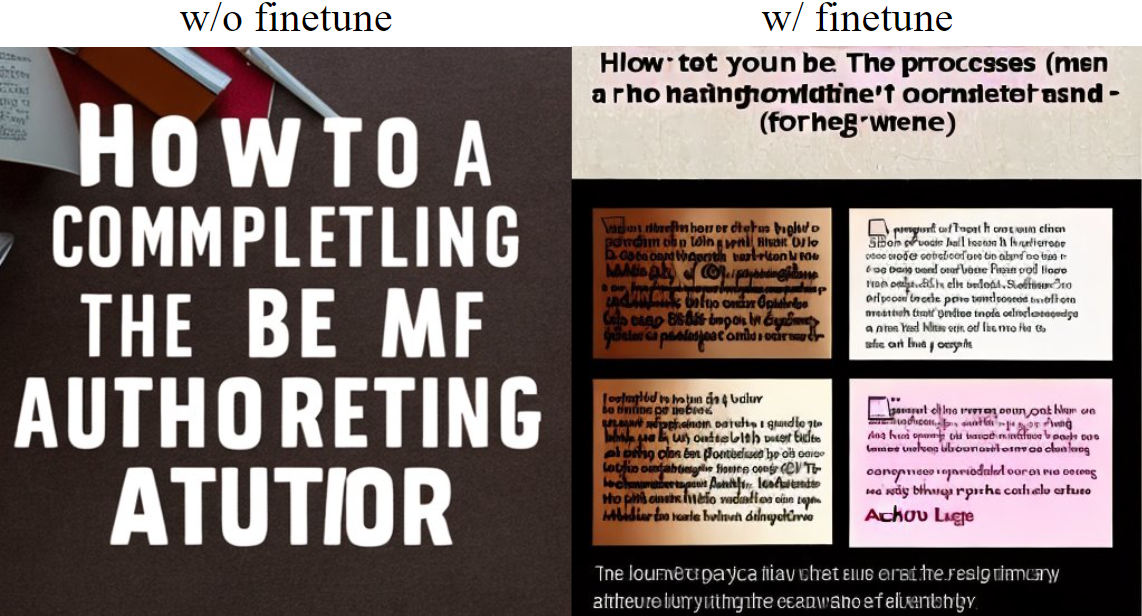}
   \caption{Example of Stable Diffusion generation with and without finetuning. The prompt is "screenshot with text: How to be an author of a paper. The process of completing a paper".}
   \label{fig:lora}
\end{figure}

Prior to finetuning, the Stable Diffusion model struggles to generate images resembling screenshots. As depicted in Figure~\ref{fig:lora}, when provided with a prompt describing screenshot content, the original Stable Diffusion model outputs an image dominated by a large text slogan. In contrast, the finetuned Stable Diffusion model successfully generates an image with a typical screen layout. Furthermore, as evidenced in Table~\ref{tab:abl} (f)(g), finetuning the base diffusion model demonstrably enhances the performance of the codec.

\subsection{Diffusion Rendering: Adaptor Level}
\label{sec:adaptor}

It is widely recognized that vanilla ControlNet \citep{Zhang2023AddingCC} is suboptimal for tackling low-level vision tasks \citep{Wang2023ExploitingDP}, as ControlNet's architecture is devised for high-level rather than low-level control. Specifically, ControlNet trains a separate feature encoder from scratch and employs residual layers to control Stable Diffusion's UNet. However, for low-level tasks, these encoder and residual layers lack the strength necessary for effective control. To improve low-level control, \citet{Wang2023ExploitingDP} introduce StableSR. which harnesses the VAE encoder of Stable Diffusion and replaces the residual layers with SPADE layers \citep{Park2019SemanticIS}. In super-resolution tasks, StableSR significantly outperforms ControlNet. As shown in Figure~\ref{fig:abl} and Table~\ref{tab:abl} (b)(c), replacing ControlNet \citep{Zhang2023AddingCC} with StableSR \citep{Wang2023ExploitingDP} improves reconstruction PSNR but reduces text accuracy. This issue arises because StableSR utilizes the Stable Diffusion's VAE (SDVAE) encoder, which is optimized for images. However, when faced glyph images, the performance of the SDVAE encoder becomes worse than ControlNet.
% \begin{figure}[thb]
%   \centering
%    \includegraphics[width=\linewidth]{fig_adap.PNG}
%    \caption{The neural network of ferature encoder in different adaptors: (a). ControlNet \citep{Zhang2023AddingCC}; (b). StableSR \citep{Wang2023ExploitingDP}; (c). Proposed. }
%    \label{fig:adap}
% \end{figure}

To address this challenge, we propose a hybrid approach that leverages the strengths of ControlNet and StableSR. Specifically, for glyph image $\bar{Z}$, we utilize only the feature encoder from ControlNet, avoiding potential losses by the SDVAE encoder. For the MLIC reconstruction $\bar{X}$, both the ControlNet feature encoder and SDVAE encoder are employed. Besides, we employ pixel shuffle \citep{Shi2016RealTimeSI} to obtain a lossless transform of $\bar{X}$. These features are concatenated to form the final conditional feature embedding before the SPADE conditioning layers (See Appendix A.2). The SDVAE features provides good representation without training, while ControlNet and pixel shuffle feature is more informative. Our evaluation in Figure~\ref{fig:abl} and Table~\ref{tab:abl} (b)(c)(d) reveals that our adaptor provides the optimal performance.

\begin{figure*}
    \centering
    \includegraphics[width=\linewidth]{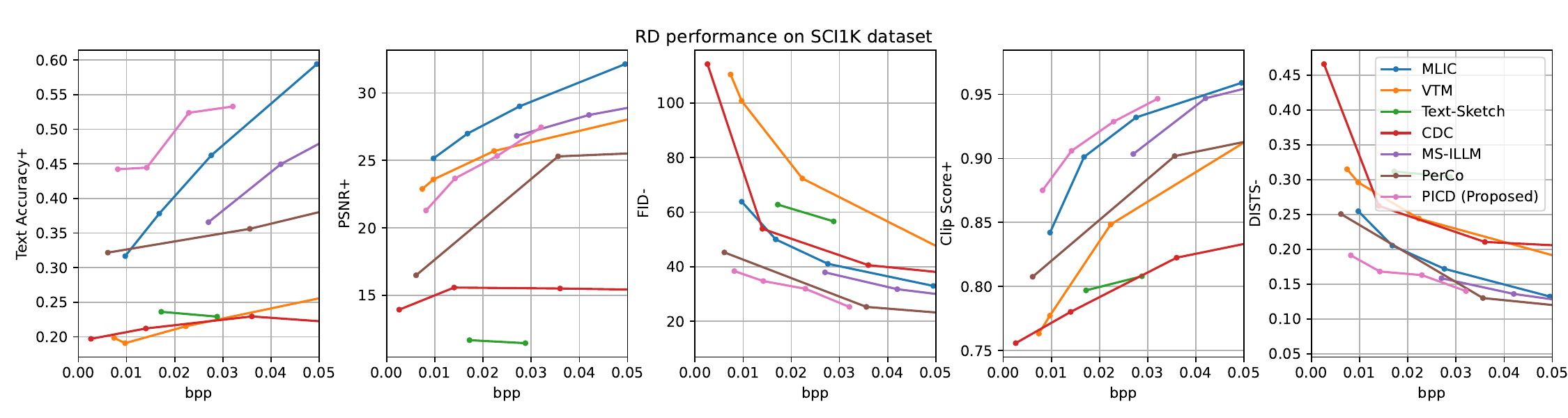}
    \includegraphics[width=\linewidth]{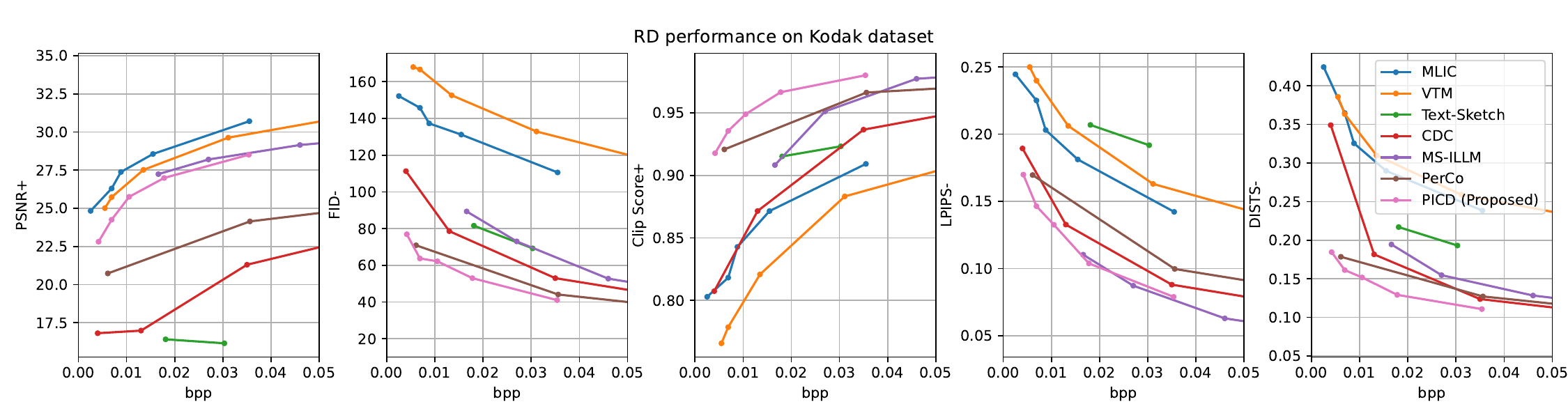}
    \caption{The rate distortion (RD) curve on screen and natural images.}
    \label{fig:rd}
\end{figure*}

\subsection{Diffusion Rendering: Instance Level}
\label{sec:ins}
To further enhance the performance of diffusion rendering, we introduce instance-level guidance during the sampling of the adapted diffusion model. In the standard DDPM \citep{ho2020denoising} process, ancestral sampling occurs from time $T$ to $0$:
\begin{gather}
    \text{DDPM Step: } X_{t-1} \sim p_{\theta}(X_{t-1}|X_t,\bar{X},\bar{Z}). \notag
\end{gather}
In practice, the conditional distribution $p_{\theta}(X_{t-1}|X_t,\bar{X},\bar{Z})$ may not be perfectly trained. To address this, instance-level guidance is applied after each DDPM step to strengthen the conditioning on $\bar{X}$ and $\bar{Z}$. Drawing on previous works in controlled generation \citep{chung2022diffusion,yu2023freedom}, we minimize the distance between the posterior mean $\mathbb{E}[X_0|X_t]$ and the conditional input. Let $X_t$ represent the current diffusion state at timestep $t$, $h(.)$ the OCR model, $f_{\theta}(.)$ the encoder, $g_{\theta}(.)$ the decoder, and $\zeta_1,\zeta_2$ hyper-parameters controlling guidance strength. We include the following guidance in each DDPM step:
\begin{align}
    \text{Guidance: } X_{t-1} &= X_{t-1} - \nabla_{X_t}\mathcal{L}(X_t, \bar{X}, \bar{Z}), \notag \\
    \text{where } \mathcal{L}(X_t, \bar{X}, \bar{Z}) &= \zeta_1 \left\| h(\mathbb{E}[X_0|X_t]) - \bar{Z} \right\| \notag \\
    &+ \zeta_2 \left\| g_{\theta}(f_{\theta}(\mathbb{E}[X_0|X_t])) - \bar{X} \right\|.
\end{align}
The first term in $\mathcal{L}(X_t,\bar{X},\bar{Z})$ ensures that the OCR output from the intermediate decoded image $\mathbb{E}[X_0|X_t]$ at time $t$ aligns with the glyph image $\bar{Z}$. Here we discard the position information and use the MSE of one-hot text information and the output logits of OCR algorithms following \citep{Lai2024LearnedIC}. The second term in $\mathcal{L}(X_t,\bar{X},\bar{Z})$ mandates that the recompressed version of the intermediate decoded image $\mathbb{E}[X_0|X_t]$ approximates the MLIC decoded image $\bar{X}$, which reduces colour drifting issue as observed by \citet{lin2023diffbir,Xu2024IdempotenceAP}.

\subsection{Simplification to a Natural Image Codec} An effective perceptual screen image codec inherently functions as an effective natural image codec, as the constraints for perceptual screen images are more stringent than those for natural images. We can simplify our PICD into a natural image codec by the following modifications: We set the glyph input $\bar{Z} = \emptyset$; We use the captions generated by BLIP \citep{Li2022BLIPBL} as text input $Z$; We train MLIC with a target mix of MSE and LPIPS; We remove screen images finetuning and the OCR loss in instance-level guidance.

\begin{table*}[thb]
\resizebox{\linewidth}{!}{
\begin{tabular}{@{}lcccccccccc@{}}
\toprule
 & \multicolumn{5}{c}{SCI1K (Screen Image)} & \multicolumn{5}{c}{SIQAD (Screen Image)} \\ \cmidrule(lr){2-6} \cmidrule(lr){7-11}
 & BD-TEXT$\uparrow$ & BD-PSNR$\uparrow$ & BD-FID$\downarrow$ & BD-CLIP$\uparrow$ & BD-DISTS$\downarrow$ & BD-TEXT$\uparrow$ & BD-PSNR$\uparrow$ & BD-FID$\downarrow$ & BD-CLIP$\uparrow$ & BD-DISTS$\downarrow$ \\ \midrule
\multicolumn{3}{@{}l@{}}{\textit{MSE Optimized Codec}}   &    &   &   &    &    &    &  \\
MLIC \citep{jiang2023mlic} (Baseline)  &  0.000  &  0.00  & 0.00   & 0.000  & 0.000  &  0.000  &  0.00  &  0.00  & 0.000  & 0.000 \\ 
VTM-SCC \citep{Bross2021OverviewOT} & -0.168 & -1.99 & 31.84 & -0.062 & 0.047 & -0.026 & 0.69 & 19.26 & -0.039 & 0.028 \\
\midrule
\multicolumn{3}{@{}l@{}}{\textit{Perceptual Optimized Codec}}   &    &   &   &    &    &    &  &  \\
% HiFiC \citep{Mentzer2020HighFidelityGI} &    &    &    &   &   & &    &    &   &  \\
Text-Sketch \citep{Lei2023TextS} & -0.135 & -14.97 & 9.98 & -0.095  & 0.087  & \underline{0.014} & -11.68 & 14.56 & -0.077 &  0.074 \\
CDC \citep{yang2022lossy} & -0.160 & -11.10 & -0.22 &  -0.091 & 0.041  & -0.090 & -7.83 & -43.25 & -0.090 &  -0.006 \\
MS-ILLM \citep{muckley2023improving} & \underline{0.025} & \textbf{-2.59} & -2.03 & -0.121 & -0.034 & -0.108 & \underline{-3.03} & -40.60   & -0.150 & -0.041 \\
PerCo \citep{Careil2023TowardsIC} & -0.057 & -5.01 & \underline{-19.90} & \underline{-0.023} & \underline{-0.035} & -0.035 & -4.46 &  \underline{-52.47} & \underline{-0.029} & \underline{-0.049} \\
% PICD-s(Proposed)  &  & & &  & &  &  & &  &  \\
PICD (Proposed)  & \textbf{0.107} & \underline{-2.97} & \textbf{-20.68} & \textbf{0.030} & \textbf{-0.050} & \textbf{0.086} & \textbf{-2.37} & \textbf{-52.93} & \textbf{0.045} & \textbf{-0.090} \\ \bottomrule 
\end{tabular}
}
\resizebox{\linewidth}{!}{
\begin{tabular}{@{}lcccccccccc@{}}
 & \multicolumn{5}{c}{Kodak (Natural Image)} & \multicolumn{5}{c}{CLIC (Natural Image)} \\ \cmidrule(lr){2-6} \cmidrule(lr){7-11}
 & BD-PSNR$\uparrow$ & BD-FID$\downarrow$ & BD-CLIP$\uparrow$ & BD-LPIPS$\downarrow$ & BD-DISTS$\downarrow$ & BD-PSNR$\uparrow$ & BD-FID$\downarrow$ & BD-CLIP$\uparrow$ & BD-LPIPS$\downarrow$ & BD-DISTS$\downarrow$ \\ \midrule
\multicolumn{3}{@{}l@{}}{\textit{MSE Optimized Codec}}   &    &   &   &    &    &    &  & \\
MLIC \citep{jiang2023mlic} (Baseline)  &   0.00  &  0.00  & 0.000   & 0.000  & 0.000  &  0.00  & 0.00   & 0.000   & 0.000 & 0.000 \\ 
VTM \citep{Bross2021OverviewOT} & -0.77 & 20.22 & -0.038 & 0.018 & 0.012 &  -1.06  &  30.19  &  -0.043  & 0.022 & 0.017 \\
\midrule
\multicolumn{3}{@{}l@{}}{\textit{Perceptual Optimized Codec}}   &    &   &   &    &    &    &  & \\
% HiFiC \citep{Mentzer2020HighFidelityGI} &    &    &    &   &   & &    &    &   &  \\
Text-Sketch \citep{Lei2023TextS} & -12.13 & -41.81 & 0.027 & 0.030 & -0.066 & -14.81 & -34.92 & \underline{0.020} & 0.070 & -0.036 \\
CDC \citep{yang2022lossy} & -10.68 &  -54.47  &  0.012  &  -0.055 & -0.114  & -11.18 &  -44.62  &  0.006  & -0.022 &  \underline{-0.099} \\
MS-ILLM \citep{muckley2023improving} & \textbf{-1.59} & -43.70 & 0.013 & \textbf{-0.069} & -0.068 & \textbf{-2.04} & -22.07 & 0.005 & \underline{-0.038} & -0.052 \\
PerCo \citep{Careil2023TowardsIC} & -6.27 & \underline{-71.29} & \underline{0.077} & -0.067 & \underline{-0.138} & -8.98 &  \underline{-48.86}  &  -0.001  &  -0.001  & -0.076\\
% PICD-s(Proposed)  &  & & &  & &  &  & &  &  \\ 
PICD (Proposed)  & \underline{-2.03} & \textbf{-74.55} & \textbf{0.084} & \underline{-0.067} & \textbf{-0.157} & \underline{-2.52} & \textbf{-61.35}  & \textbf{0.057} & \textbf{-0.043} & \textbf{-0.134} \\ \bottomrule
\end{tabular}
}
\caption{Quantitative results on screen and natural images. \textbf{Bold} and \underline{Underline}: Best and second best performance in perceptual codec.}
\label{tab:quant}
\end{table*}

\section{Experimental Results}
\subsection{Experimental Setup}
All experiments were conducted using a computer equipped with an A100 GPU, CUDA version 12.0, and PyTorch version 2.1.0. (See Appendix B.1 for details).

\textbf{Datasets} For screen content experiments, the models were trained using the WebUI dataset \citep{Wu2023WebUIAD} containing 400,000 screen images. Subsequently, the models were evaluated on the \textbf{SCI1K} \citep{Yang2021ImplicitTN} and \textbf{SIQAD} \citep{Yang2015PerceptualQA}. For natural image experiments, model training was conducted using the OpenImages dataset \citep{Kuznetsova2018TheOI}, and evaluations were performed using the \textbf{Kodak} \citep{Kodakdataset} and \textbf{CLIC} \citep{Clicdataset}.

\textbf{Evaluation Metrics} To assess text accuracy of screen content, the Jaccard similarity index is employed. For assessing quality of images, we use the Fréchet Inception Distance (\textbf{FID}) \citep{Binkowski2018DemystifyingMG}, Learned Perceptual Image Patch Similarity ( \textbf{LPIPS}) \citep{zhang2018unreasonable}, \textbf{CLIP similarity} \citep{Radford2021LearningTV}, Deep Image Structure and Texture Similarity (\textbf{DISTS}) \citep{Ding2020ImageQA} and Peak signal-to-noise ratio (\textbf{PSNR}). To enable comparison between codecs operating at different bitrates, we calculate the Bjontegaard (BD)  metrics \citep{bjontegaard2001calculation}, with a bits-per-pixel (bpp) ranging from $0.005-0.05$.

\textbf{Baselines} For baselines, we select several state-of-the-art perceptual image codecs, including \textbf{Text-Sketch} \citep{Lei2023TextS}, \textbf{CDC} \citep{yang2022lossy},  \textbf{MS-ILLM} \citep{muckley2023improving}, and an opensourced implementation of \textbf{PerCo} \citep{Careil2023TowardsIC}, namely \textbf{PerCo (SD)} \citep{Korber2024PerCoO}. Additionally, we include mean squared error (MSE) optimized codecs in our comparison, such as \textbf{MLIC} \citep{jiang2023mlic} and \textbf{VTM} \citep{bross2021developments}. We acknowledge that there are many other very competitive baselines \citep{mao2024extreme,Jia2024GenerativeLC,Li2024TowardEI,Lee2024NeuralIC}, while we can only include several most related works.  

\subsection{Main Results}
\begin{figure*}[thb]
\centering
\includegraphics[width=\linewidth]{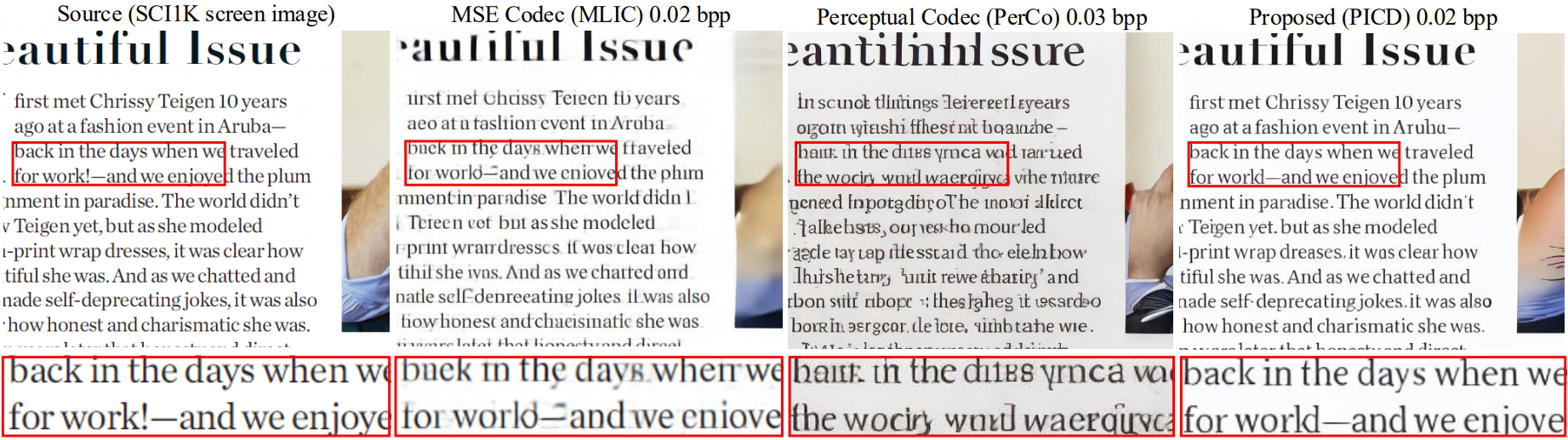}
\includegraphics[width=\linewidth]{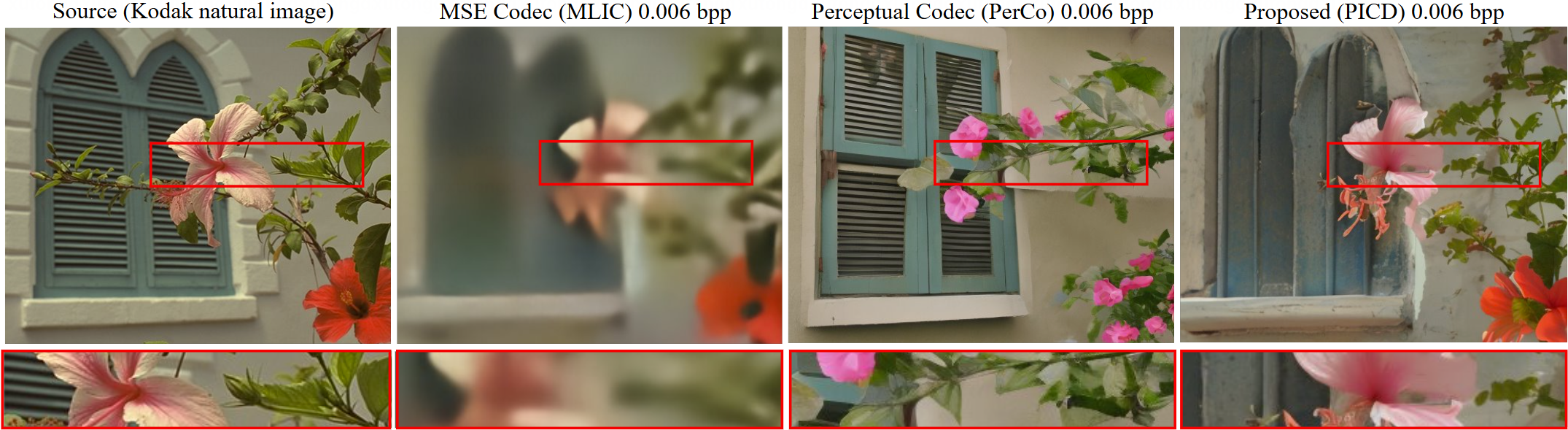}
\captionof{figure}{Qualitative results on screen and natural images.}
\label{fig:qual}
\end{figure*}

\textbf{Results on Screen Images} As illustrated in Table~\ref{tab:quant}, PICD excels in achieving both high text accuracy and superior perceptual quality. Notably, PICD achieves the highest text accuracy and the lowest Fréchet Inception Distance (FID) among all the assessed methods. While other perceptual codecs also manage to achieve low FID, they fall short in text accuracy. This discrepancy is visually depicted in Figures~\ref{fig:cov} and~\ref{fig:qual}, where perceptual codecs such as PerCo produce visually sharp reconstructions but fail to accurately reconstruct text content. In contrast, our PICD succeeds in maintaining high visual quality alongside accurate text reconstruction. 

\textbf{Results on Natural Images} Furthermore, as evidenced in Table~\ref{tab:quant} and Figures~\ref{fig:cov} and~\ref{fig:qual}, our PICD performs effectively on natural image compression as well. Specifically, PICD achieves the lowest FID. This result underscores PICD's competency as a highly effective perceptual codec for natural images.

% \usepackage{booktabs}
\begin{table}[htb]
\includegraphics[width=\linewidth]{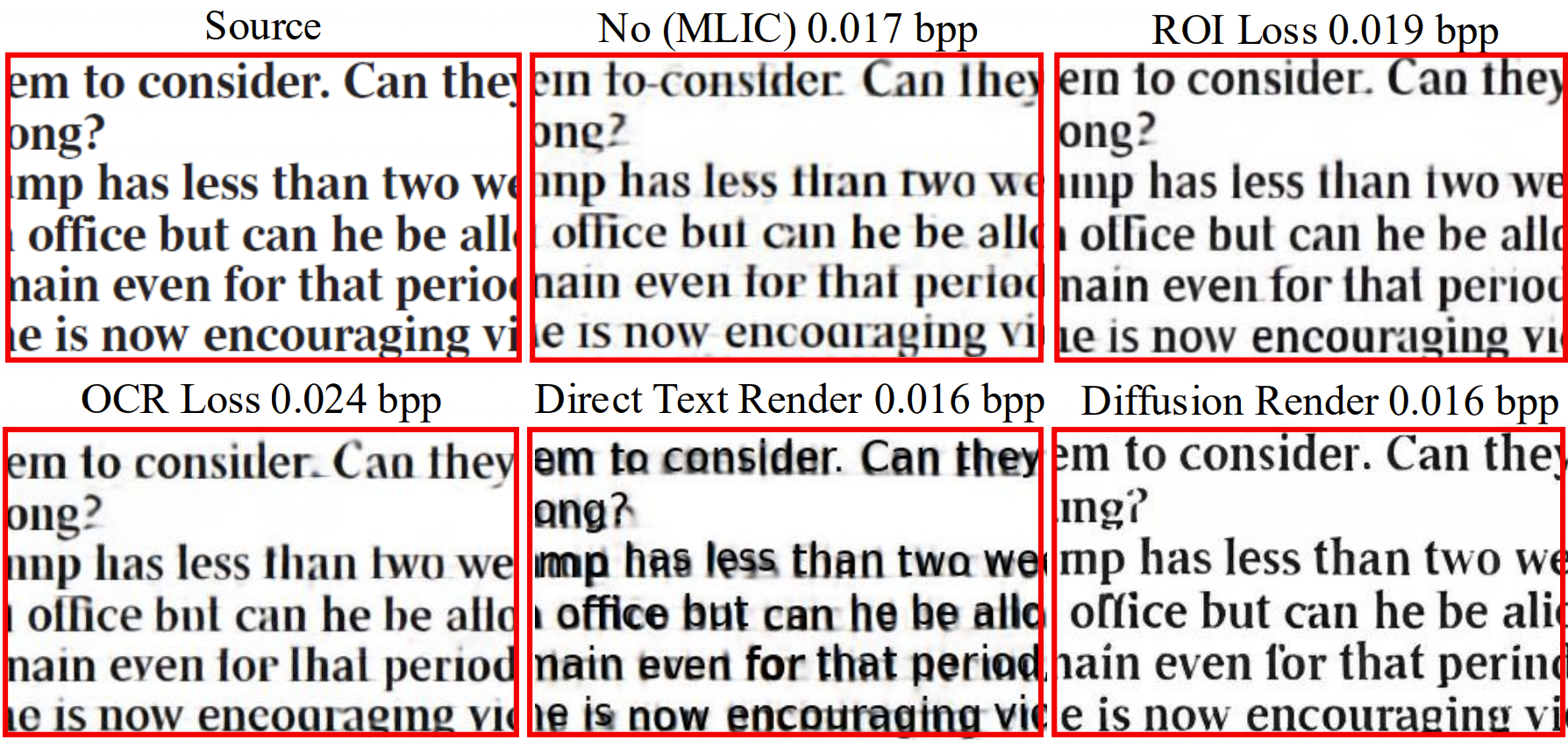}
\captionof{figure}{Visual results of different text coding tools.}
\label{fig:abl2}
\centering
\resizebox{\linewidth}{!}{
\begin{tabular}{@{}lcccc@{}}
\toprule
Text Coding Tools & bpp$\downarrow$ & Text Acc$\uparrow$ & FID$\downarrow$ & CLIP$\uparrow$ \\ \midrule
No (MLIC \citep{jiang2023mlic}) & 0.017 & 0.221 & 57.25 & 0.8269 \\
ROI Loss \citep{Och2023ImprovedSC,Heris2023MultiTaskLF,zhou2024enhanced}            & 0.019 & 0.250 & 54.51 & 0.8280 \\
OCR Loss \citep{Lai2024LearnedIC} & 0.024 & 0.251 & 51.33 & 0.8344 \\
Direct Text Render \citep{Mitrica2019VeryLB,Tang2022TSASCCTS} & 0.016 & 0.463 & 52.20 & 0.8785 \\
Diffusion Render (Proposed) & 0.016 & 0.445 & 34.77 & 0.9059 \\ \bottomrule
\end{tabular}
}
\captionof{table}{Ablation studies on the means of preserving text content.}
\label{tab:abl2}
\end{table}

\subsection{Ablation Studies}
\textbf{Effects of Diffusion Rendering and Three Level Improvements} To gain a deeper understanding of the contributions made by various components of our PICD, we conducted ablation studies focusing on screen content. The results, presented in Table~\ref{tab:abl}, indicate the following: 1) Incorporating glyph images is crucial for effective text reconstruction, as evidenced by the comparison between setups (a) and (b); 2) Our proposed adaptor demonstrates superior effectiveness compared to alternatives like ControlNet and StableSR, as seen in the comparison of (d) against (b) and (c); 3) Improvements at both the instance level and domain level further enhance performance, as demonstrated by the comparisons of (f) versus (d), and (g) versus (f).

\textbf{Alternative Text Tools for Diffusion Rendering} Previous literature on screen content codecs has developed various other text tools to improve text accuracy, including ROI loss \citep{Och2023ImprovedSC,Heris2023MultiTaskLF,zhou2024enhanced}, OCR loss \citep{Lai2024LearnedIC}, and direct text rendering \citep{Mitrica2019VeryLB,Tang2022TSASCCTS}. To better understand the performance of our diffusion rendering relative to these methods, we compared them using MLIC as the base codec. As shown in Table~\ref{tab:abl2}, all the proposed approaches successfully improve text accuracy compared to the base MLIC \citep{jiang2023mlic}. Among all the methods, direct text rendering shows slightly better text accuracy than diffusion rendering, but it suffers from a significantly worse FID and CLIP similarity. Overall, only our diffusion rendering achieves both high text accuracy and high visual quality simultaneously.

\begin{table}[htb]
\centering
\resizebox{0.8\linewidth}{!}{
\begin{tabular}{@{}lccc@{}}
\toprule
                & Training        & Encoding  & Decoding     \\ \midrule
Text-Sketch \citep{Lei2023TextS}    & 0               & $\ge$120s & $\approx$10s \\
CDC \citep{yang2022lossy}            & $\approx$3 days & $\le$0.1s & $\approx$10s \\
MS-ILLM \citep{muckley2023improving}        & $\approx$7 days & $\le$0.1s & $\le$0.1s    \\
PerCo \citep{Careil2023TowardsIC}          & $\approx$7 days & $\le$0.1s & $\approx$10s \\
PICD-(d) (Proposed) & $\approx$1 day  & $\le$0.1s & $\approx$10s \\
PICD (Proposed) & $\approx$1 day  & $\le$0.1s & $\approx$30s \\ \bottomrule
\end{tabular}
}
\captionof{table}{Temporal complexity of different approaches.}
\label{tab:comp}
\end{table}

\textbf{Temporal Complexity} In Table~\ref{tab:comp}, we compare the temporal complexity of different methods on 512$\times$512 images. Due to the efficient design of PCID, the training time of PICD is significantly lower than other approaches, which takes around a day on a single A100 GPU. Without instance level guidance, our PICD-(d) has comparable encoding and decoding complexity as other diffusion based approaches. With instance level guidance, our PICD takes around 30s to decode. Additionally, PICD has around the
same model size, same peak memory usage and twice FLOPS compared with PerCo \citep{Careil2023TowardsIC}.

\section{Related Works}
\textbf{Perceptual Image Compression} The majority of perceptual image codecs leverage conditional generative models to achieve exceptional perceptual quality at very low bitrates, using a conditional GAN \citep{Rippel2017RealTimeAI,tschannen2018deep,Mentzer2020HighFidelityGI,Agustsson2022MultiRealismIC,muckley2023improving,Jia2024GenerativeLC} or diffusion model \citep{yang2022lossy,hoogeboom2023high,Careil2023TowardsIC,Lei2023TextS,Ma2024CorrectingDP,Relic2024LossyIC}. Among them, the text guided codec are particularity related to our work \citep{Careil2023TowardsIC,Lei2023TextS,Lee2024NeuralIC}. Our simplified natural codec can be seen as a variant of those works, which are post-processing based compressed image restoration \citep{hoogeboom2023high,Wang2023SinSRDI}. However, most perceptual image codecs fail when applied to screen content.

\textbf{Screen Content Compression} Conversely, many coding tools are developed to enhance text accuracy in screen content. Some methods allocate additional bitrate to text regions \citep{ Heris2023MultiTaskLF}. Other techniques enforce constraints on Optical Character Recognition (OCR) results between the source and reconstructed images during codec training \citep{Lai2024LearnedIC}. In an approach akin to ours, \citet{Mitrica2019VeryLB,Tang2022TSASCCTS} also encode text losslessly, while they render text directly on the decoded image. While these methods significantly enhance text accuracy in screen content compression, they are not perceptual codecs and thus fail to produce visually pleasing results at low bitrates. On the other hand, our PICD achieves both high text accuracy and superior perceptual quality simultaneously.

\section{Discussion \& Conclusion}
One limitation of the PICD is its decoding speed. The instance-level guidance \citep{chung2022diffusion} requires a decoding time of approximately 30 seconds, which is 3 times slower than other diffusion codecs. Faster guidance methods \citep{yang2024guidance} could be employed, while we defer this aspect to future research.

In conclusion, we present the PICD, a versatile perceptual image codec which excels for both screen and natural images. The PICD builds on a pre-trained diffusion model by incorporating conditions across domain, adaptor, and instance levels. The PICD not only achieves high text accuracy and perceptual quality for screen content but also proves to be an effective perceptual codec for natural images.
\newpage
\maketitlesupplementary
\appendix

\section{Implementation Details}
\subsection{Neural Network Architecture of Text-conditioned MLIC}
\begin{figure}[htb]
  \centering
   \includegraphics[width=\linewidth]{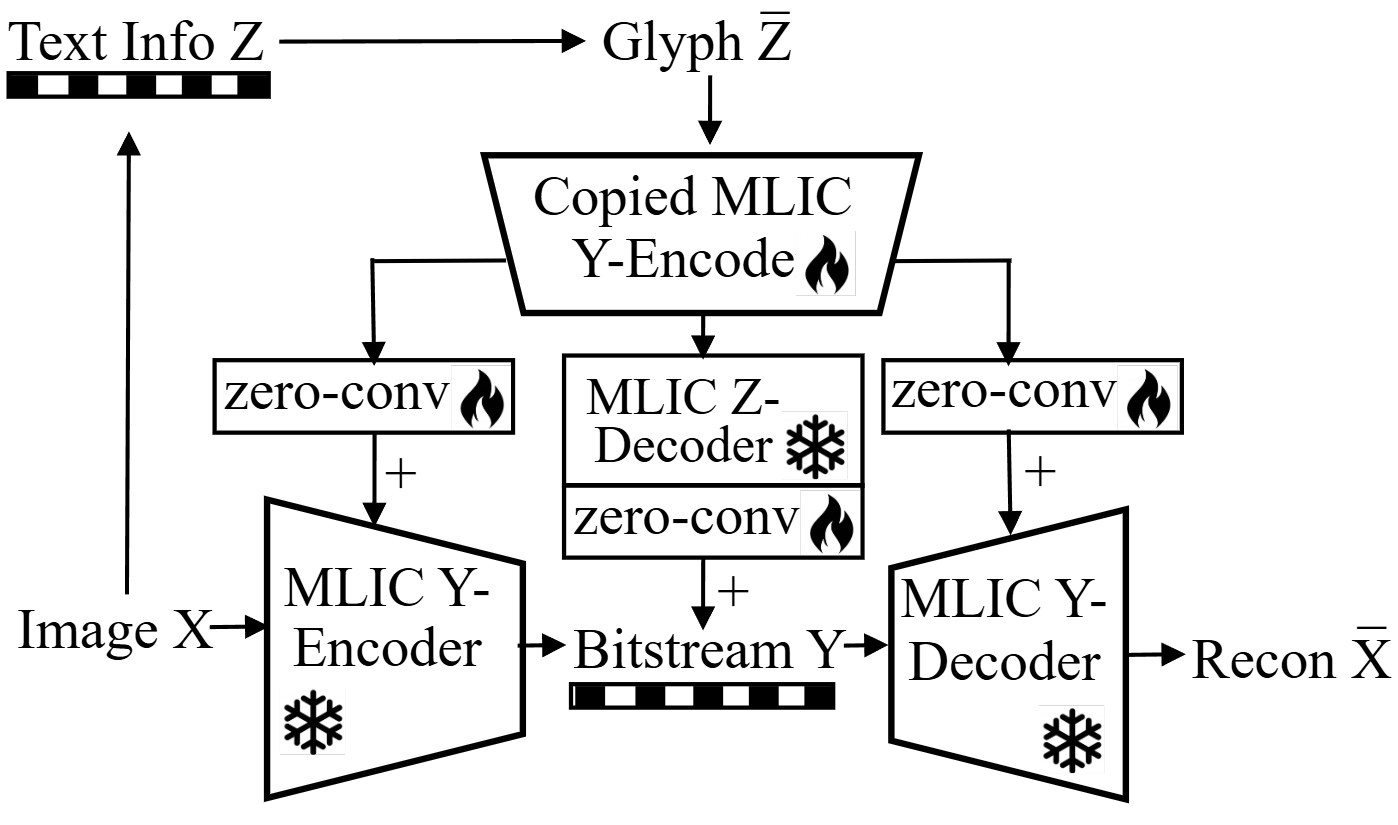}
   \caption{The neural network architecture of text-conditioned MLIC.}
   \label{fig:mlic}
\end{figure}

In Figure~\ref{fig:mlic}, we illustrate the neural network architecture of text conditioned MLIC model.
\subsection{Neural Network Architecture of Proposed Adaptor}
\begin{figure}[htb]
  \centering
   \includegraphics[width=\linewidth]{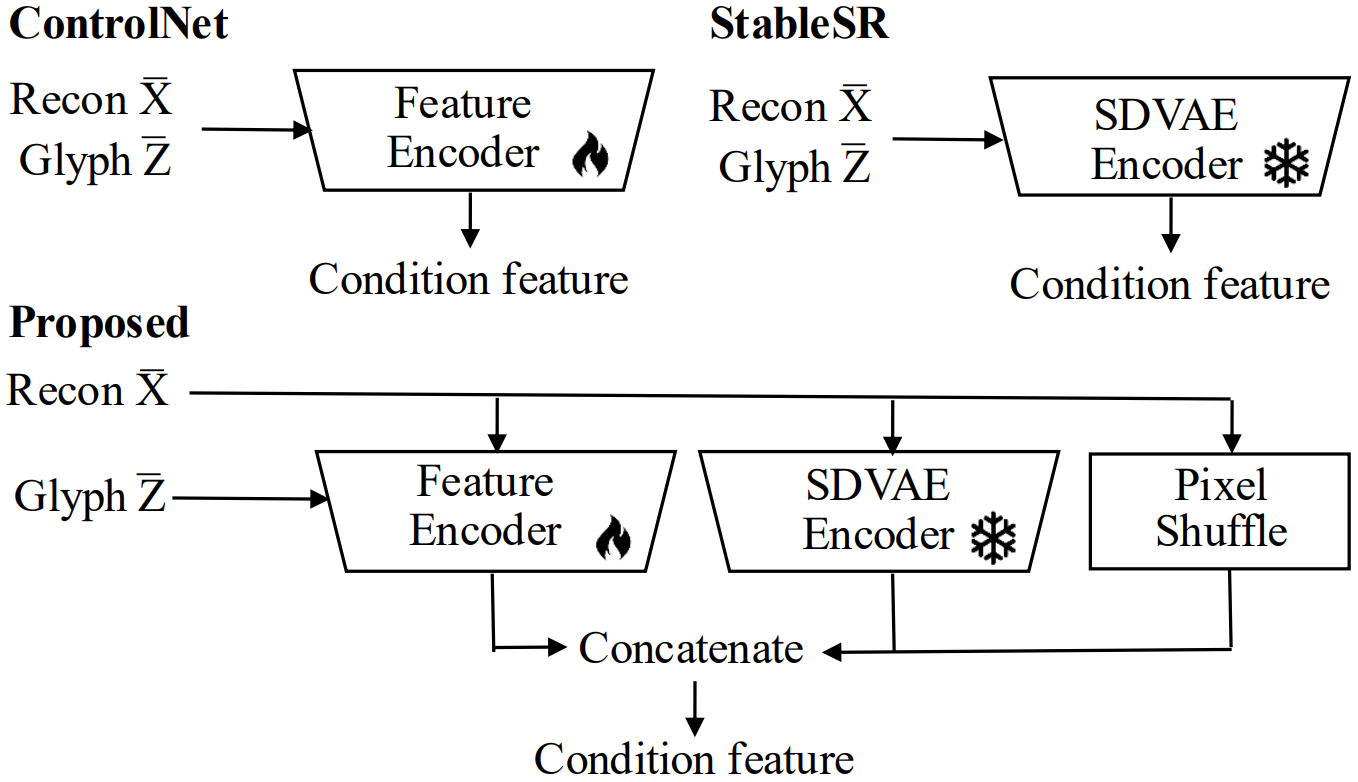}
   \caption{The neural network architecture of the proposed adaptor.}
   \label{fig:adap}
\end{figure}

In Figure~\ref{fig:adap}, we illustrate the adaptor's neural network architecture of vanilla ControlNet \citep{Zhang2023AddingCC}, StableSR \citep{Wang2023ExploitingDP} and our proposed approach.
\subsection{Instance Level Guidance}
To implement instance level guidance, we first need to obtain $\mathbb{E}[X_0|X_t, y]$ using Tweedie's formula following \citet{chung2022diffusion}:
\begin{gather}
    \mathbb{E}[X_0|X_t, y] = \frac{1}{\sqrt{\bar{\alpha}_t}}(X_t + (1-\bar{\alpha}_t)s_{\theta}(X_t,t,y)),
\end{gather}
where $s_{\theta}(.,.,.)$ is the trained score estimator of diffusion model.

The instance level guidance is composed of OCR guidance and codec guidance. The codec guidance is straightforward and details can be found in \citet{Xu2024IdempotenceAP}. While the OCR guidance is not that straightforward.

We adopt Tesseract OCR engine \citep{Smith2007AnOO} to extract text from images, following \citet{Tang2022TSASCCTS}. However, this OCR engine is not differentiable. And we can not use it in instance level OCR guidance. To solve this problem, we alternatively adopt the neural network based OCR engine named PARSeq \citet{Bautista2022SceneTR}, which is adopted in \citet{Lai2024LearnedIC}.

Next, we use the bounding box information in $Z$ to cut the source image $\mathbb{E}[X_0|X_t, y]$. Then, those slice of images are feed into PARSeq. PARSeq produces the logits, which is further compared with the true text content in $Z$ (weighted by $\zeta_1$ in Section 3.5) as guidance for diffusion model.

\subsection{Hyper-parameters of Diffusion Rendering}
In Table~\ref{tab:hyp}, we show the hyper-parameters used for diffusion rendering.
\begin{table}[htb]
\centering
\begin{tabular}{@{}lc@{}}
\toprule
      & Diffusion rendering hyper-parameter. \\ \midrule
SCI1K & $T=250, \zeta_1=0.25,\zeta_2=1e-4, \omega=0.0 $ \\
SIQAD & $T=250, \zeta_1=0.25,\zeta_2=1e-4, \omega=0.0 $ \\
Kodak & $T=500, \zeta_1=0.25,\zeta_2=0.0, \omega=3.0 $ \\
CLIC  & $T=500, \zeta_1=0.25,\zeta_2=0.0, \omega=3.0 $ \\ \bottomrule
\end{tabular}
\caption{Diffusion rendering related hyper-parameters.}
\label{tab:hyp}
\end{table}

\section{Additional Experimental Results}
\subsection{Additional Experimental Setup}
All the experiments are conducted on a computer with 1 A100 GPU. For the domain level finetuning, we train the LoRA augumented Stable Diffusion 2.0 model with batchsize 64 and 10,000 steps of gradient ascent. We use a learning rate of 1e-4 and a LoRA with rank 256. The training costs around 2 days. For the adaptor training, we adopt a batchsize 64 and 5,000 steps of gradient ascent with learning rate 1e-4 and batchsize 64. The training cost around 1 day. Note that the domain level finetuning only happens once. While for each bitrate, we need to train a different adaptor.

\subsection{Additional Quantitive Results}
\begin{figure*}
    \centering
    \includegraphics[width=\linewidth]{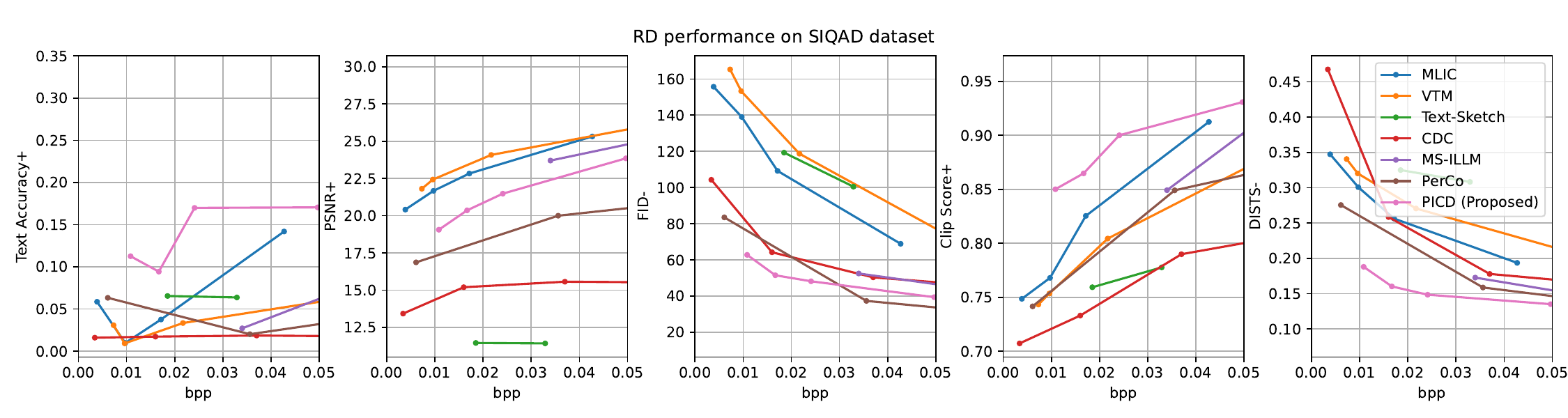}
    \includegraphics[width=\linewidth]{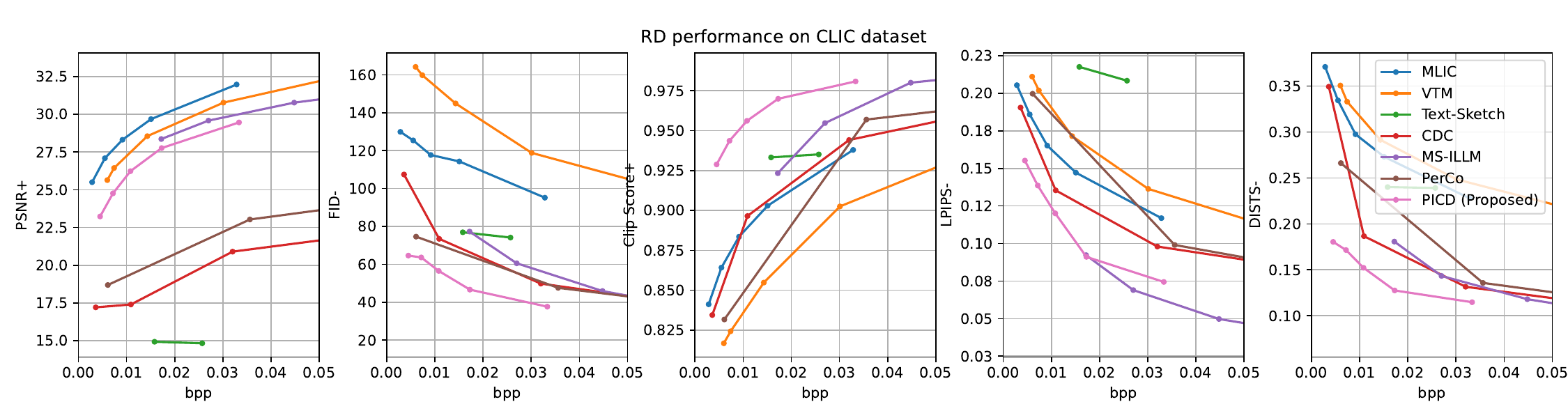}
    \caption{The rate distortion (RD) curve on screen and natural images.}
    \label{fig:rd2}
\end{figure*}

For RD performance, we also evaluate the LPIPS metric for screen contents, which is shown in Table~\ref{tab:quant2}. And in Figure~\ref{fig:rd2}, we present the RD curve on SIQAD and CLIC dataset.

\begin{table}[thb]
\resizebox{\linewidth}{!}{
\begin{tabular}{@{}lcc@{}}
\toprule
 & SCI1K (Screen) & SIQAD (Screen) \\ \cmidrule(lr){2-2} \cmidrule(lr){3-3}
 & BD-LPIPS$\downarrow$ & BD-LPIPS$\downarrow$ \\ \midrule
\multicolumn{3}{@{}l@{}}{\textit{MSE Optimized Codec}} \\
MLIC \citep{jiang2023mlic} (Baseline) & 0.000 & 0.000 \\ 
VTM-SCC \citep{Bross2021OverviewOT} & 0.055 &  0.021\\
\midrule
\multicolumn{3}{@{}l@{}}{\textit{Perceptual Optimized Codec}} \\
Text-Sketch \citep{Lei2023TextS} & 0.135 & 0.087 \\
CDC \citep{yang2022lossy} & 0.100 & 0.024 \\
MS-ILLM \citep{muckley2023improving} & \textbf{-0.023} & \textbf{-0.082} \\
PerCo \citep{Careil2023TowardsIC} & 0.001 & -0.070 \\
% PICD-s(Proposed)  &  & & &  & &  &  & &  &  \\
PICD (Proposed)  & \underline{-0.005} & \underline{-0.080} \\ \bottomrule 
\end{tabular}
}
\caption{LPIPS results on screen images. \textbf{Bold} and \underline{Underline}: Best and second best performance in perceptual codec.}
\label{tab:quant2}
\end{table}

\subsection{Additional Qualitative Results}
\begin{figure*}[thb]
\centering
\includegraphics[width=\linewidth]{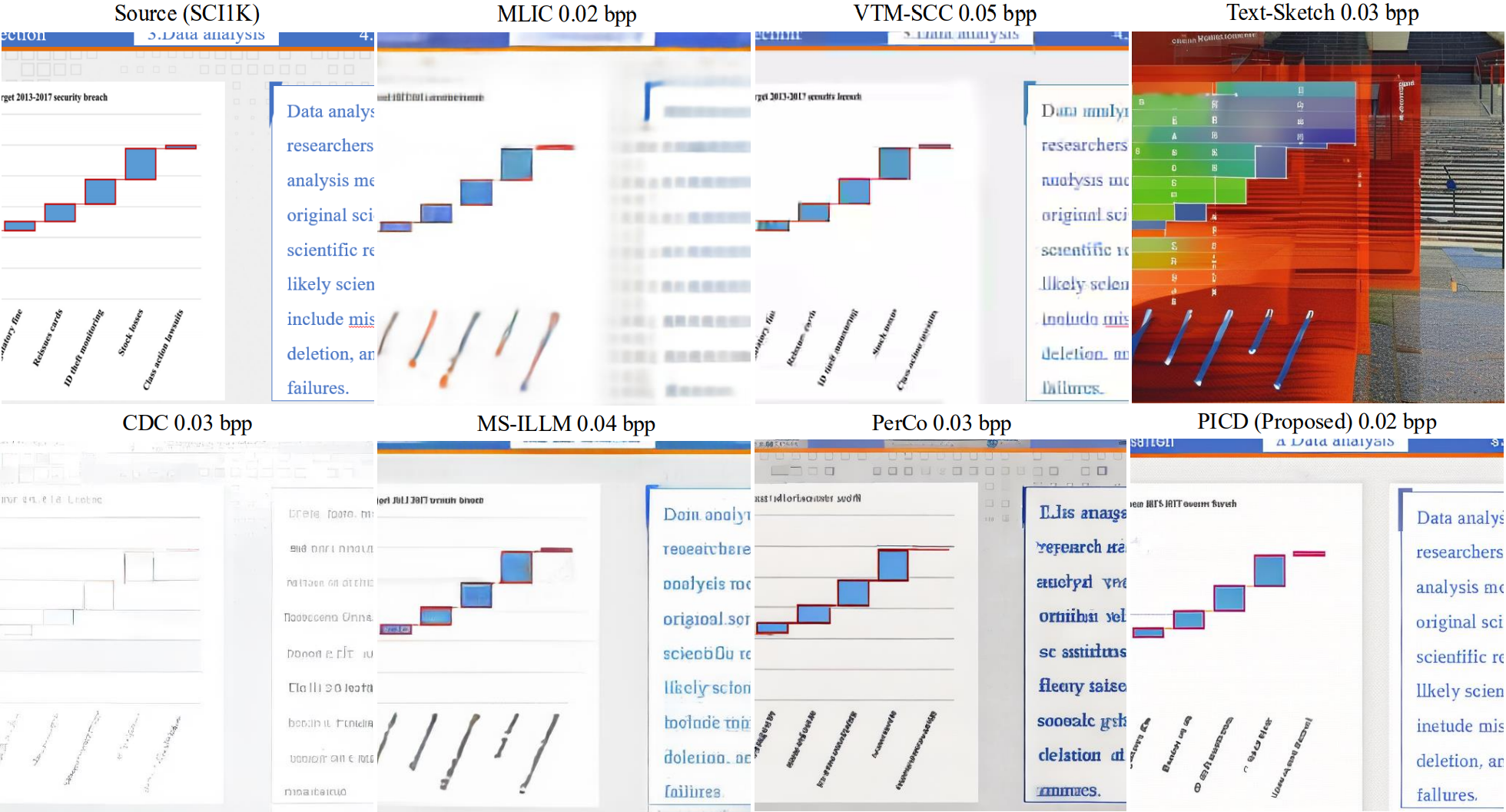}
\includegraphics[width=\linewidth]{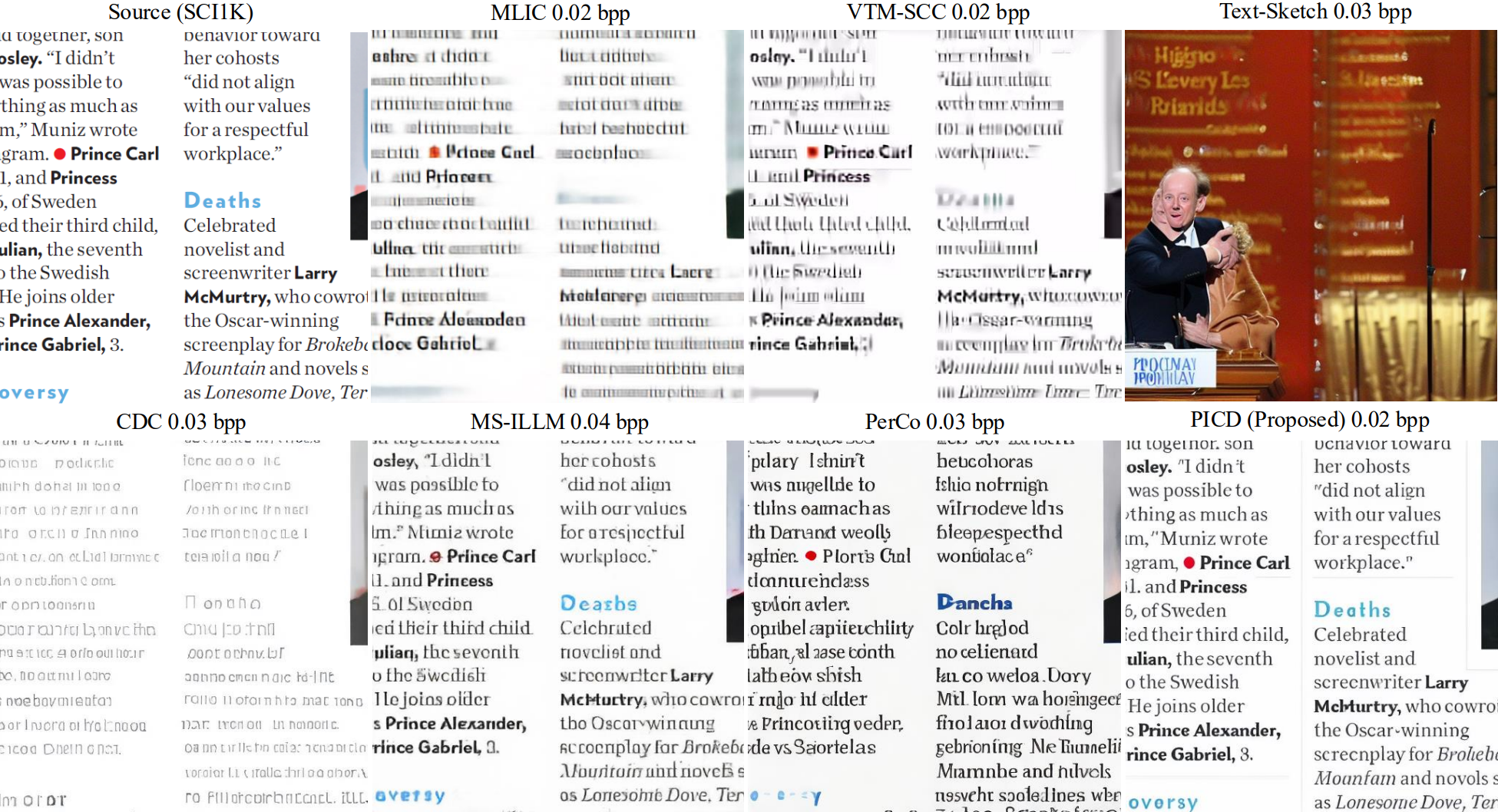}
\captionof{figure}{Qualitative results on screen images.}
\label{fig:qual3}
\end{figure*}

\begin{figure*}[thb]
\centering
\includegraphics[width=\linewidth]{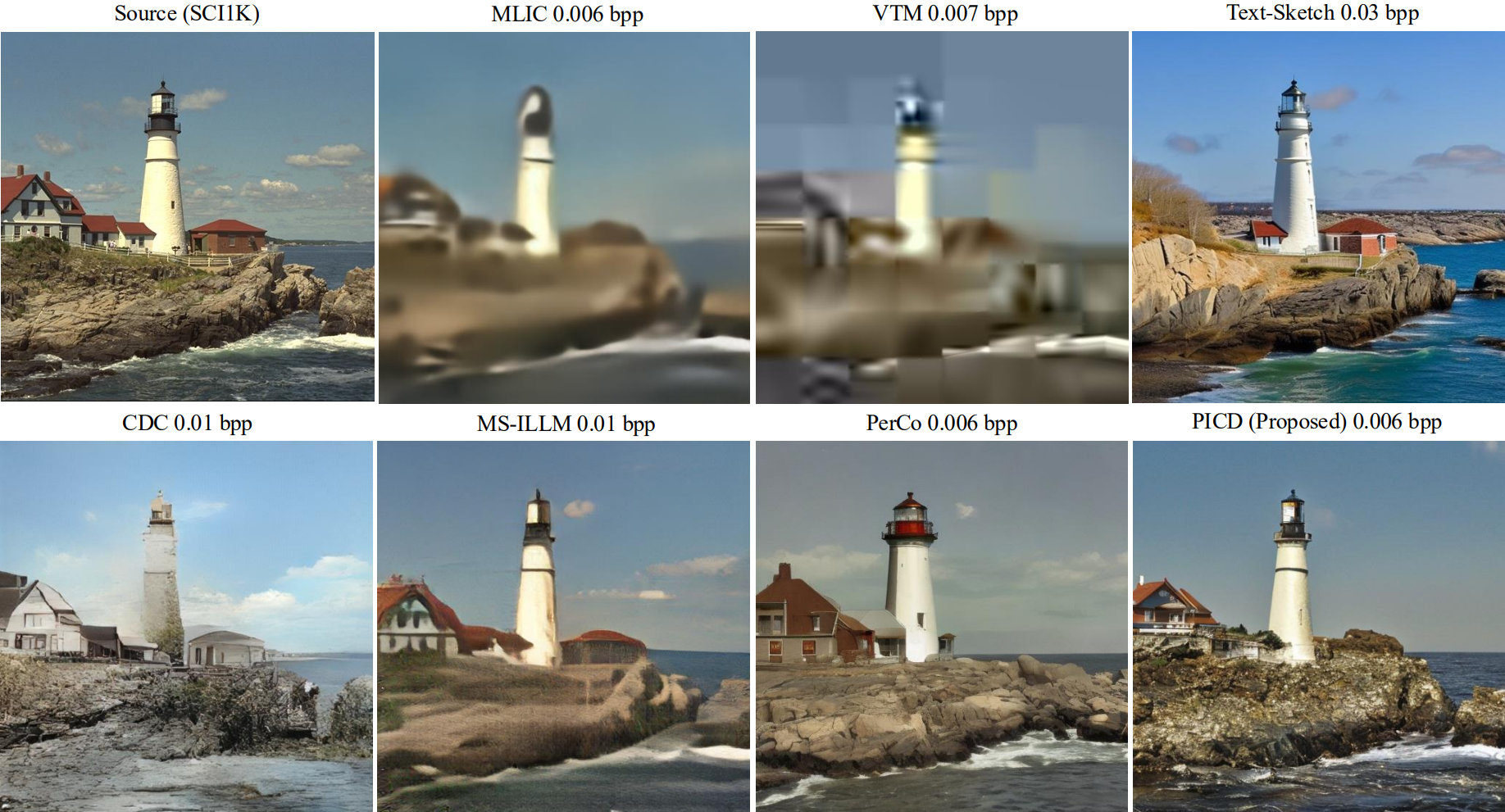}
\includegraphics[width=\linewidth]{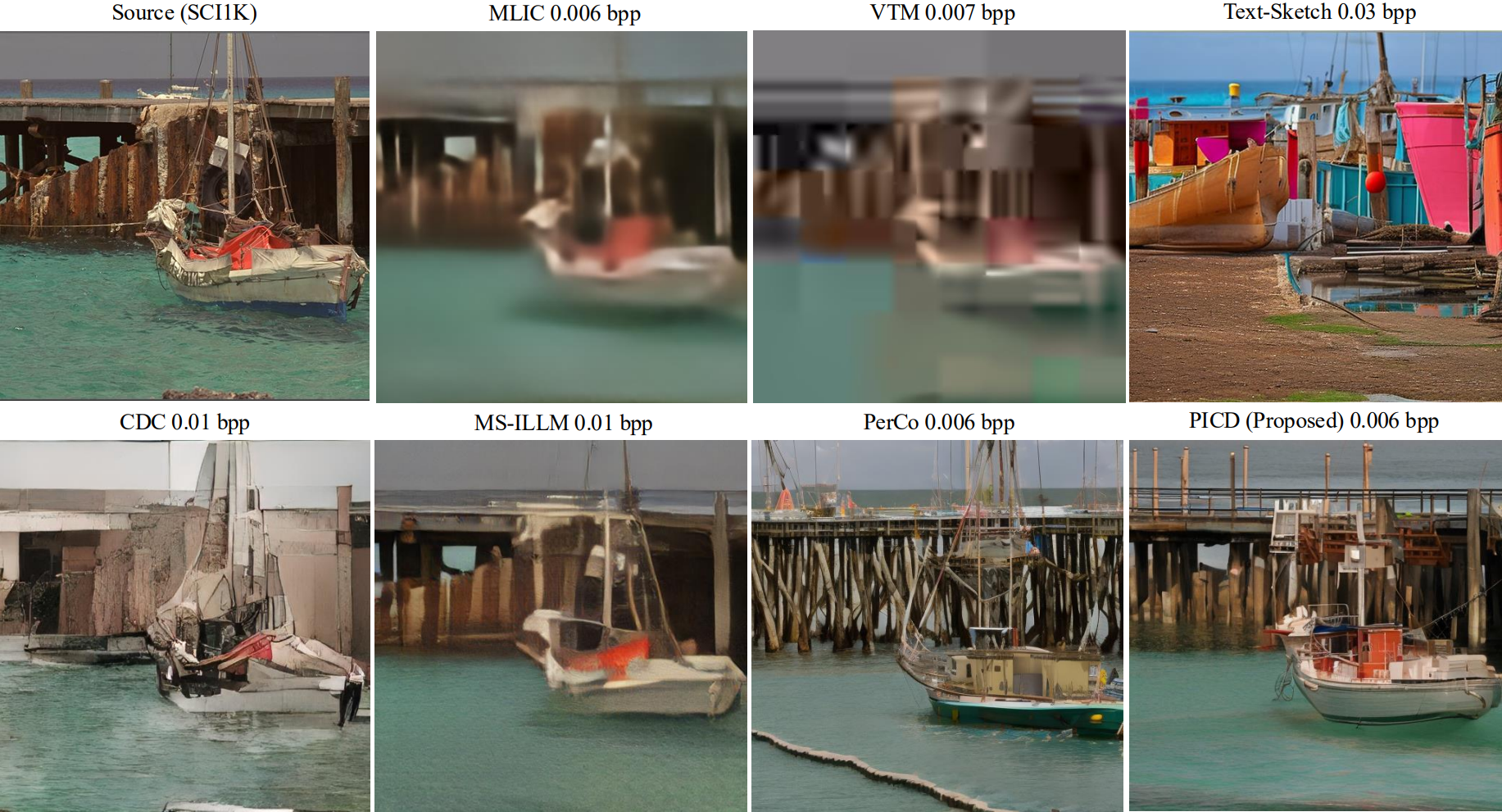}
\captionof{figure}{Qualitative results on natural images.}
\label{fig:qual4}
\end{figure*}

We present more qualitative results in Figure~\ref{fig:qual3}-\ref{fig:qual4}.

\subsection{Additional Ablation Studies}

\begin{table}
\centering
\begin{tabular}{@{}lcccc@{}}
\toprule
CFG & FID$\downarrow$ & PSNR$\uparrow$ & CLIP$\uparrow$ & LPIPS$\downarrow$ \\ \midrule
0.0 & 71.37 & 24.47 & 0.9247 & 0.1498 \\
3.0 & 63.76 & 24.25 & 0.9356 & 0.1464 \\
5.0 & 68.00 & 23.74 & 0.9274 & 0.1555 \\
7.0 & 70.14 & 23.67 & 0.9213 & 0.1575 \\ \bottomrule
\end{tabular}
\caption{Ablation study on classifier-free guidance (CFG) for natural images.}
\label{tab:ablcfg}
\end{table}
% \usepackage{booktabs}

\textbf{Classifier-free Guidance} Additionally, in the context of PICD for natural image compression, we discovered the significant importance of classifier-free guidance (CFG). Table~\ref{tab:ablcfg} illustrates that varying levels of CFG markedly affect the FID and PSNR. Through empirical evaluation, we determined that a CFG value of 3.0 optimizes results, yielding the best FID, CLIP similarity, and LPIPS. This finding is consistent with observations reported by \citet{Careil2023TowardsIC}.

\subsection{MS-SSIM as Perceptual Metric} In both our setting and other papers (ILLM), MS-SSIM aligns more with PSNR than visual quality. In our case, for SCI1K dataset, the BD-MS-SSIM is: MLIC (0.01) $>$ VTM (0.00) $>$ ILLM (-0.003) $>$ PICD (-0.006). We are reluctant to use MS-SSIM as perceptual metric, as it is obviously not aligned with visual quality. In CLIC codec competition [50], the best human rated codec has almost worst MS-SSIM. We will emphasis that MS-SSIM is not a perceptual metric, and include those results.

\subsection{Failure Case} 
\begin{figure}[htb]
    \centering
    \includegraphics[width=\linewidth]{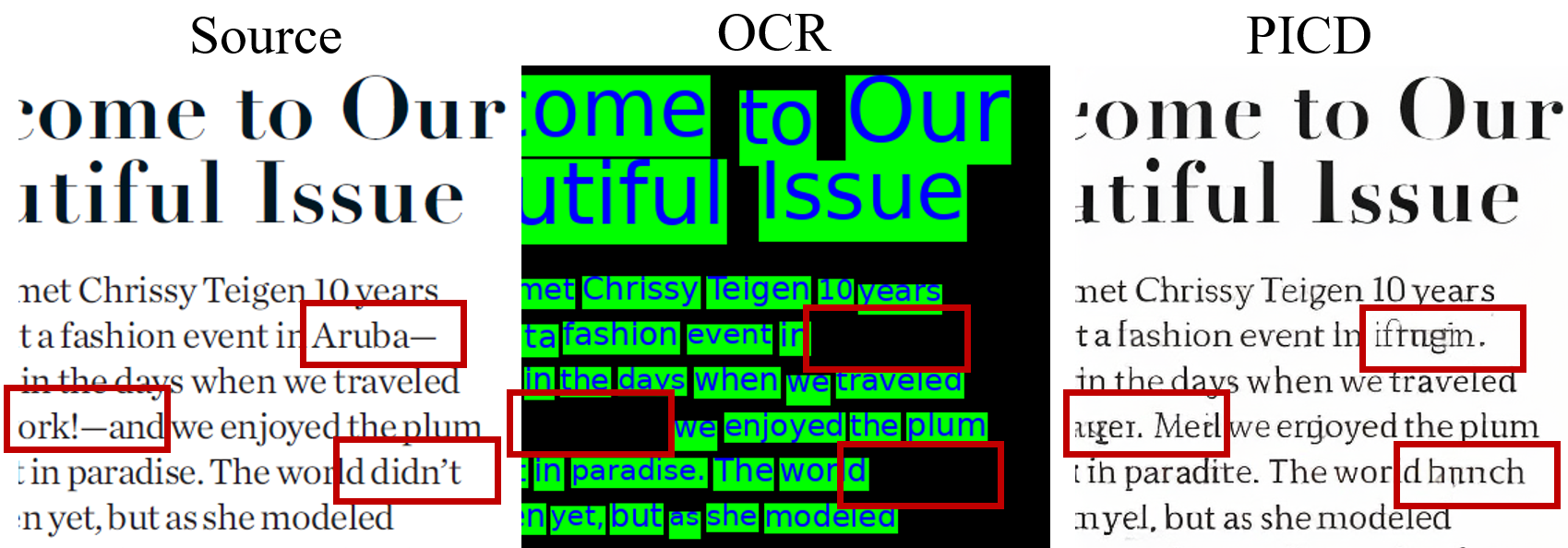}
    \caption{An example of OCR failure.}
    \label{fig:reb}
\end{figure}

Our text rendering fails if the OCR algorithm fails. Typically, an OCR failure brings distortion and mis-rendering of text content. A visual example is shown in Fig.~\ref{fig:reb}.

\clearpage

{
    \small
    \bibliographystyle{ieeenat_fullname}
    \bibliography{main}
}

\end{document}